\documentclass[10pt, a4paper]{article}
\usepackage{lrec}
\usepackage{multibib}
\newcites{languageresource}{Language Resources}
\usepackage{graphicx}
\usepackage{tabularx}
\usepackage{booktabs}
\usepackage{soul}
\usepackage{longtable}
\usepackage{pdfpages}

\usepackage{epstopdf}
\usepackage[utf8]{inputenc}
\usepackage{todonotes}
\usepackage{hyperref}
\usepackage{xstring}
\usepackage{csvsimple}
\usepackage{array,ragged2e}
\newcolumntype{P}[1]{>{\RaggedRight\arraybackslash}p{#1}}

\usepackage{color}

\title{CATs are Fuzzy PETs:  A Corpus and Analysis of Potentially Euphemistic Terms}

\name{Martha Gavidia, Patrick Lee, Anna Feldman, Jing Peng}

\address{Montclair State University\\
         Montclair, NJ, USA \\
         \{gavidiam1, leep6, feldmana, pengj\}@montclair.edu\\}

\abstract{
Euphemisms have not received much attention in natural language processing, despite being an important element of polite and figurative language. Euphemisms prove to be a difficult topic, not only because they are subject to language change, but also because humans may not agree on what is a euphemism and what is not. Nevertheless, the first step to tackling the issue is to collect and analyze examples of euphemisms. We present a corpus of potentially euphemistic terms (PETs) along with example texts from the GloWbE corpus. Additionally, we present a subcorpus of texts where these PETs are not being used euphemistically, which may be useful for future applications. We also discuss the results of multiple analyses run on the corpus. Firstly, we find that sentiment analysis on the euphemistic texts supports that PETs generally decrease negative and offensive sentiment. Secondly, we observe cases of disagreement in an annotation task, where humans are asked to label PETs as euphemistic or not in a subset of our corpus text examples. We attribute the disagreement to a variety of potential reasons, including if the PET was a commonly accepted term (CAT). 
 \\\newline  }

\begin{document}

\maketitleabstract

\section{Introduction}
Euphemisms are mild or indirect expressions that are used in place of more unpleasant or offensive ones. They can be used in generally polite conversation (e.g., \emph{passed away} instead of \emph{died}), or even as a way to try and hide the truth \cite{rababah2014translatability} by minimizing a threat and downplaying situations in order to create a favorable image \cite{karam2011truths}; for example, saying \emph{armed conflict} instead of \emph{war}. For this paper, we consider a variety of euphemistic expressions that provide alternatives to more direct meanings. These alternatives allow us to avoid potential awkwardness or offensiveness when discussing sensitive or taboo topics such as death, sexual activity, employment, bodily functions, politics, physical/mental attributes, etc. Because of their figurative nature, euphemisms can be ambiguous, either because of human subjectivity, or because euphemistic words and phrases may also be used literally (e.g., \emph{between jobs}). As such, working with euphemisms in natural language processing is not straightforward.

In this paper, we present an expanded list of common euphemisms which we will refer to as \emph{PETs}, or \emph{Potentially Euphemistic Terms}, as well as a corpus of euphemistic and literal usages of said PETs from web-based text data.  To the best of our knowledge, there are no existing corpora of English sentences containing euphemisms.  We hope the development of these new resources help build upon current NLP applications surrounding euphemisms, particularly in providing context differences to disambiguate these terms. 

The rest of the paper is as follows: Section 2 reviews previous work done on the language of politeness and how it relates to euphemisms as well as current computational approaches on euphemism recognition, detection and generation. Section 3 gives details on how we compiled our corpus including information on our list of PETs, source text, and sentence extraction and selection.
In Section 4, we describe our corpus and provide examples of euphemistic and literal usages of PETs within our corpora. Section 5 discusses experimental results of sentiment analysis with a roBERTa-base model \cite{liu2019roberta,barbieri2020tweeteval}; we theorize that there is a shift in sentiment and offensiveness of PETs vs. their literal meanings in the same context since the usage of euphemisms makes our speech less emotionally charged. We also describe an annotation task and offer possible explanations as to why euphemisms are ambiguous. Section 6 finally concludes with a discussion about future work.

\section{Related Work}
\subsection{Euphemisms and Politeness}
Euphemisms are related to the language of politeness, e.g.,  \cite{danescu2013computational,rababah2014translatability}, which plays a role in applications involving dialogue and social interactions in different contexts, including political discourse or doctor-patient interactions.  Politeness has been a central concern in pragmatic theory \cite{grice1975logic,leech1983pragmatics,lakoff1973language,lakoff1979stylistic,brown1987politeness} because we can learn about language, culture and society through the language of politeness. 

\newcite{danescu2013computational} propose a computational framework for identifying linguistic aspects of politeness and use their framework to study the relationship between  politeness and social power.
Politeness is learned within our communities and daily social interactions, so it is natural that when we communicate on the internet some of those features are carried over with us as is seen in certain aspects of online social communication, including forums and message boards.  This assumption drives our investigation into the use of euphemisms in web based data as a politeness marker.

\newcite{madaan2020politeness} introduce a task of politeness transfer which involves converting non-polite sentences to polite sentences while preserving the meaning. \newcite{madaan2020politeness}  adopt the data-driven approach to politeness proposed by \newcite{danescu2013computational}  Additionally, they create a corpus of 1.39 million instances automatically labeled for politeness. They propose a \emph{tag} and \emph{generate} pipeline that identifies stylistic attributes and generates a sentence in the target style while preserving most of the source content. While this work is concerned with style transfer rather than euphemisms, we find this work relevant, especially, for the euphemism generation task.\\
Additionally, \newcite{chaves2021should} discuss the growing popularity of chatbots and conduct a survey on eleven social characteristics a chatbot can have that benefit human-chatbot interactions; manners is defined as one of them.  They refer to manners as the ability of a chatbot to manifest polite behavior and conversational habits \cite{chaves2021should,morrissey2013realness}.  The adoption of speech acts such as greetings, apologies, and closings \cite{jain2018evaluating} and minimizing impositions \cite{tallyn2018ethnobot,toxtli2018understanding} are a few of the ways in which chatbots currently manifest politeness \cite{chaves2021should}.  We find this relevant to the future studies in euphemism generation as a way to manifest politeness. 

\subsection{Previous Computational Work on Euphemisms}
There is not much computational work on recognizing and interpreting euphemisms. The most directly related work is by \newcite{magu2018determining}, \newcite{felt2020recognizing}, \newcite{kapron2021diachronic}, \newcite{zhu2021self} and \newcite{zhu2021euphemistic}.

\newcite{felt2020recognizing} present the first effort to recognize  \emph{x-phemisms}, euphemisms and dysphemisms (derogatory terms), using NLP. They identify near-synonym phrases for three topics (FIRING, LYING, and STEALING) using a weakly supervised bootstrapping algorithm for semantic lexicon induction \cite{thelen2002bootstrapping}. Next, they classify phrases as euphemistic, dysphemistic, or neutral using lexical sentiment cues and contextual sentiment analysis. Additionally, they contribute a gold-standard dataset of human x-phemism judgements. \newcite{thelen2002bootstrapping} show that sentiment connotation and affective polarity are useful for identifying x-phemisms, but not sufficient and while the performance of \newcite{felt2020recognizing}'s system is relatively low and the range of topics is very narrow, this work is certainly inspiring further investigations. 

\newcite{zhu2021self} define two tasks:  1) euphemism detection (based on the input keywords, produce a list of candidate euphemisms) 2) euphemism identification (take the list of candidate euphemisms produced in (1) and output an interpretation). They approach the task as an unsupervised fill in the mask problem and use a masked language model twice: 1) to filter the masked sentences and 2) generate the euphemism candidates from the masked sentences.  For euphemism identification (=interpretation), \newcite{zhu2021self} extract phrases from a base corpus, then use word embeddings' similarities to filter out ones that are associated with a seed list of euphemisms, then finally use a masked language model SpanBERT to rank the euphemistic candidates.
Their system outperforms all the baselines including \newcite{felt2020recognizing}. The technical innovation of this work relies on the idea of self-supervision \cite{zhu2021self}, a form of unsupervised learning  where the data itself provides the supervision. While the approach appears promising, it has a number of limitations. Like \newcite{felt2020recognizing}'s system, \newcite{zhu2021self}  rely on a set of predefined terms (topics such as drugs, weapons, and sexuality). The system is not capable to discover new contexts in which euphemisms are used. In addition, \newcite{zhu2021self} treat euphemisms as mere substitutions. In this respect their work is similar to \newcite{magu2018determining}, who also treat code words as euphemisms. Their euphemism detection system has a specific purpose: content moderation, so in a sense, they are just detecting "inappropriate topics/contexts", not necessarily euphemisms. Still, this work opens new avenues for euphemism detection and interpretation. It is the first work that uses BERT for processing euphemisms. \newcite{zhu2021euphemistic} still define the task of euphemism detection vaguely, but they improve upon \newcite{zhu2021self}'s approach by adding an automatic paraphraser.

Lastly, \newcite{kapron2021diachronic} investigate gender differences in language under the assumption that women use euphemisms more than men. Their work debunks the assumption.  However, through this investigation, they assemble a list of 106 euphemism-taboo pairs to analyze their relative use through time by each gender in the corpora which served as a valuable resource in our PETs dataset creation which we discuss next. 

\section{Corpus Creation}
Using a list of PETs, we extract sentences from a base corpus \cite{davies2015expanding}  and manually annotate each as either euphemistic or non-euphemistic (literal).  We then select 1,382 euphemistic sentences and 583 additional sentences in which select PETs were also found to be used literally.  For the PETs dataset and corpus see: \url{https://github.com/marsgav/euphemism_project}.

\subsection{Potentially Euphemistic Terms (PETs)} 
We compile a collection of 184 PETs from several sources including euphemism dictionaries, English websites designed for second language learners, online articles highlighting some of the top most common euphemisms as well as our own linguistic knowledge of euphemisms \cite{kapron2021diachronic,rawson1981dictionary,holder2008dictionary,englishclub,joneseuphsarticle,silvereuphsarticle,woelfeleuphsarticle,gormandyeuphsarticle,dictionary1989oed,hereemaeuphsarticle,oconnereuphsarticle,martin1991negro}. We chose these different sources to make sure we cover a variety of taboo topics, but also to keep up with the common euphemisms as of January 2022. Euphemisms are constantly being created and removed; the "euphemism treadmill" describes how euphemisms can sometimes become offensive over time and thus lose their euphemism status \cite{pinker2003blank}. While a definitive list of euphemisms can never be created, we aim to cover a variety of euphemisms relating to death, sexual activity, employment, bodily functions, politics, physical/mental attributes, substances, and other miscellaneous taboo topics.

\subsection{The GloWbE Corpus}
The Corpus of Global Web-Based English (GloWbE) corpus \cite{davies2015expanding} contains 1.9 billion words of text from twenty different English speaking countries.  Its inclusion of 20 different dialects of English makes it an optimal source for examining euphemisms since euphemisms are cultural and geographical. For this reason, our extracted sentences are derived from only a portion of the \emph{US dialect} of English text contained within GloWbE. 

\subsection{Sentence Extraction and Selection}
We use spaCy's PhraseMatcher \cite{spacy2} to identify rows from our raw text data which contain terms from our pre-defined list.  

\subsubsection{PhraseMatcher for Sentence Extraction}
PhraseMatcher allows users to efficiently match large terminology lists in texts by an exact token match. For this reason, we included several morphological variations for some of our PETs in our search.  The terminology list fed into PhraseMatcher contained 284 total terms to account for all 184 PETs plus their added variations.  Examples of these variations, as illustrated in Table \ref{tbl:morphvar}, include (1) taking pronoun changes into account, (2) pluralization, and (3) tense changes.
\begin{table}[htb]
\begin{center}
\begin{tabularx}{\columnwidth}{l|X}

       \hline
       \textbf{PET} & \textbf{Variations}\\
       [0.5ex]\hline
       lose your lunch & lose/lost my lunch, lose/lost her lunch, lose/lost his lunch, lose/lost their lunch, lose/lost our lunch\\
       senior citizen & senior citizens\\
       lay off & laid off, laying off\\
       \hline

\end{tabularx}

\caption{Examples of morphological variations included in PET list}
\label{tbl:morphvar}
\end{center}

\end{table}

\subsubsection{Sentence Selection}
PhraseMatcher yielded an output of over 5,500 rows of text containing some of our target PETs. Every row had a different amount of text so we preprocessed our text to include the sentence containing the target PET as well as 1-3 surrounding sentence for added context.  We then manually annotated every row as either '1' -- euphemistic or '0' -- non-euphemistic.  
Given the ambiguous nature of euphemisms, we had a disproportionate amount of non-euphemistic texts vs. euphemistic texts.  In an attempt to create a balanced corpus, we selected a maximum of 30 sentences for every PET found with \emph{PhraseMatcher} that was used in a euphemistic sense.  Results of this yielded a total of 1,382 \emph{euphemistic} sentences spanning 129 different PETs.  

We follow the same methodology to select a maximum of 30 sentences for the PETs that were also found to be used in a non-euphemistic sense.  This sub-corpus contains 583 \emph{non-euphemistic} (literal) sentences spanning only 58 out of 129 total PETs.  In other words, our corpus contains 71 PETs that were always found in a euphemistic sense and 58 PETs that were found with both a euphemistic and literal sense.  See Appendix A and B for both lists of PETs.  We include this sub-corpus as it may contain valuable insights that can be gathered from the context differences surrounding the euphemistic and literal usages of PETs.

\section{Corpus Details}
We combine our euphemistic and literal examples into one corpus of 1,965 total sentences. We label the 71 PETs that were always used euphemistically as \textit{always euphs} and the 58 found with both usages as \textit{sometimes euphs}.  The euphemistic and literal usages of \textit{sometimes euphs} can be easily filtered within the corpus. The average word count per example is 65 words, and the average character count is 373.  These details are summarized in Table \ref{tbl:corpusDesc}.

Additionally, we examine the taboo or sensitive topics that our euphemistic PETs cover; these are displayed in Table \ref{tbl:sensitive} along with example PETs.  Table \ref{tbl:euphandlit} contains some examples of \textit {sometimes euphs} PETs being used both euphemistically and literally;  here, the role of context in distinguishing euphemistic vs literal usages may be observed.
\begin{table}[htb]
\begin{center}
\begin{tabularx}{\columnwidth}{lc}
\hline
\multicolumn{2}{l}{\textbf{Euphemism Corpus}} \\ [0.5ex]\hline
Total Sentences              & 1, 965         \\
Euphemistic Sentences        & 1, 382         \\
\textit{with always euphs} PET& 777           \\
\textit{with sometimes euphs} PET& 605        \\
Literal Sentences            & 583            \\
Avg. Word Count              & 65 words       \\
Avg. Char. Count             & 373 char.      \\
Total PETs                   & 129            \\
\textit{always euphs}        & \textit{71}    \\
\textit{sometimes euphs}     & \textit{58}    \\ \hline
\end{tabularx}
\caption{Euphemism Corpus Description}
\label{tbl:corpusDesc}
\end{center}
\end{table}

\begin{table*}[!h]
\begin{center}
\begin{tabular}{lll}
\hline
\textbf{Taboo/Sensitive Topic} & \textbf{Count} & \textbf{PET Examples}                                       \\ [0.5ex]\hline
death                        & 214   & pass away, late, put to sleep                      \\
sexual   activity            & 68    & make love, sex worker, sleep   with                \\
employment                   & 246   & lay off, dismissed, downsize                       \\
bodily   functions           & 51    & accident, pass gass, time of the month                       \\
politics                     & 279   & freedom fighter, correctional   facility, pro-life \\
physical/mental   attributes & 387   & aging, overweight, disabled                        \\
substances                   & 115   & weed, substance abuser, getting   clean            \\ \hline
\end{tabular}
\caption{Sensitive/Taboo Topics with PET examples}
\label{tbl:sensitive}
\end{center}
\end{table*}

\begin{table*}[!h]
\begin{center}
\begin{tabular}{llp{10cm}}
\textbf{PET}& \textbf{Label}& \textbf{Example}\\ \hline
\multicolumn{1}{l|}{weed}& \textit{euphemistic} & You will want to stop off at the   medical marijuana dispensary for a supply of fireworks, alcohol, personal   weaponry and dope-\textless weed\textgreater. Then, fill a glass or pop a top   or load a bong or whatever one does, to get along these days.\\
\multicolumn{1}{l|}{}& \textit{literal}& In some ways, cultivating for   \textless weed \textgreater control is almost a lost art. Herbicides seemed   to work so well for so long that many farmers abandoned mechanical means of   control.\\ \hline
\multicolumn{1}{l|}{disabled} & \textit{euphemistic} & No no no no. I'm in the same   situation-- \textless disabled \textgreater, chronic pain, artist, no   "visible disability" (even when I'm in my chair), and nobody   understands that it takes us longer to do *everything*. I'm honestly   surprised you even humored your neighbor this far!\\
\multicolumn{1}{l|}{}& \textit{literal}& They claim there is no network   or storage capability in these machines, clearly this is not true. These   features may be \textless disabled \textgreater or only available to   administrators who service the equipment, but in any event the TSA @ @ @ @ @   @ @ @ @ @ problems. As to the veterans out there who work for the TSA, I   share your frustration \\ \hline
\multicolumn{1}{l|}{between jobs} & \textit{euphemistic} & I would still donate food and clothing for people in need but at least I would know that it was my choice and it was being used for it's intended purpose. I applied for temporary assistance when I was \textless between jobs \textgreater for a month to support my family. We had no savings or income and we were denied because I had made too much money the previous year.\\
\multicolumn{1}{l|}{}& \textit{literal}& The more new people you meet, the more your chances of finding out about a great job increases. Then if you hear back from multiple places, you'll have choices and who wouldn't want to be able to choose \textless between jobs \textgreater rather than grasping at the first one that comes along. \\ 

\end{tabular}
\caption{Euphemistic and Literal Usages of PETs}
\end{center}
\label{tbl:euphandlit}
\end{table*}

\section{Experiments}
\subsection{roBERTa for Sentiment and Offensive Ratings}
Since euphemisms are used with the aim to be polite, like \newcite{felt2020recognizing}, we hypothesize that the sentiment of a sentence containing a euphemism should generally be more positive and less offensive \cite{bakhriddionova2021needs}. To investigate this, we performed a sentiment analysis on our corpus, in which sentiment and offensiveness scores were computed for each text sample, and then re-computed after substituting each phrase with its literal meaning. An example substitution is shown below:
\newline

\emph{Just from my personal observations, among \textbf{low-income} kids, those with a strong home life tend to do better.} 
\newline
\centerline{$\downarrow$}
\emph{Just from my personal observations, among \textbf{poor} kids, those with a strong home life tend to do better. }
\newline

The sentiment scores were computed using a roBERTa-based model, which was trained on Tweets (which is suitable for our examples' informal text), fine-tuned for sentiment analysis and offensive language identification, and evaluated using the TweetEval framework \cite{liu2019roberta,barbieri2020tweeteval}. The scores before and after substitution are compared using relative change, since each score is a probability of a classification label, rather than an absolute score (and should therefore be considered relative to that particular text). Table \ref{tbl:offensivenessChange} shows the average percent changes after replacement of the literal meaning.
\begin{table}[!h]
\begin{center}
\begin{tabularx}{\columnwidth}{c c c} 
 \hline
 \textbf{Model} & \textbf{Label} & \textbf{\% Change} \\ [0.5ex] 
 \hline
 Sentiment & Neutral & -2.6\% \\
  & Positive & -11.3\% \\
  & Negative & 54.6\% \\
 \hline
 Offensiveness & Not-Offensive & -6.6\% \\
  & Offensive & 30.0\% \\ [1ex] 
 \hline
\end{tabularx}
\end{center}
\caption{Changes in sentiment/offensiveness scores after replacement of euphemism}
\label{tbl:offensivenessChange}
\end{table}

The results indicate that the use of a euphemism, as opposed to its literal meaning, affects sentiment scores. In particular, negative and offensive scores increase noticeably after substitution, which supports the assumption that euphemism softens language \cite{bakhriddionova2021needs}. Additionally, the sentiment scores were grouped by PET and averaged, which shows the average sentiment changes per PET (see Appendix C). These results could be significant for future work involving euphemism detection using sentiment.

\subsection{Corpus Annotation Task}
Because we know that euphemisms can be interpreted differently, we decided to let language experts (graduate students of NLP at Montclair) examine what their perceived interpretations were for our selected PETs given both euphemistic and literal context.  For this final portion of our paper, we analyze their interpretations.

\subsubsection{Task Instructions}
Annotators were given a sample of 500 sentences in which the target PET was contained within $\langle$ $\rangle$.  Without supplying the annotators with the literal meanings, they were asked to follow our annotation model and enter a 1 if they considered the sentence to be euphemistic and a 0 if they considered it to be non-euphemistic given the target PET.  For every instance they were asked to provide their interpretations as well so that we could evaluate whether these PETs were in fact common enough to evoke similar interpretations.  A confidence score was also requested on a scale of 1-3 to test how confident they each were of their interpretation. Appendix D includes the task instructions given to the annotators.

\subsubsection{Inter-rater Agreement}
Recognizing and agreeing on whether a term is a euphemism or not can present some challenges given that euphemisms are ambiguous.  We were curious to examine whether the inter-rater reliability scores between our own annotations and those of our language experts reflected this ambiguity. We evaluated our observed agreement, and calculated Krippendorf's alpha to test reliability.  

\subsubsection{Observed Agreement}

We examined the observed agreement between ourselves and each individual annotator as well as the agreement between different pairs of annotators.  This is simply a measure of how frequently a pair of annotators agreed on a label.  

\subsubsection{Krippendorf's alpha}
To test inter-rater reliability we use a freely available macro written for SPSS and SAS to calculate Krippendorff's alpha \cite{hayes2007answering}.  Krippendorf's alpha, described in \newcite{krippendorff2011computing}, is a reliability coefficient which measures the agreement among any number of annotators where the general form for $\alpha$ is: \[\alpha = 1 - \frac{D_o}{D_e}\] $D_o$ being the observed disagreement among the values assigned to the units of analysis and $D_e$ being the disagreement one would expect when the coding of units is attributable to chance rather than to the properties of these units.  \cite{hayes2007answering} describe the two reliability scale points for Krippendorf alpha as 1.000 for perfect reliability and 0.000 for the absence of reliability and say that these two points enable an index to be interpreted as the degree to which the data can be relied on in subsequent analyses.
\begin{table}[!h]
\begin{center}
\begin{tabularx}{\columnwidth}{ll}
\hline
\multicolumn{2}{l}{\textbf{Inter-rater Reliability Results}} \\ [0.5ex]\hline
Annotators                              & 5          \\
Cases                               & 500        \\
Decisions                           & 2500       \\
Average observed percent   agreement  & 71.74\%    \\
Krippendorf's alpha                    & 0.415 \\ \hline
\end{tabularx}
\end{center}
\caption{Inter-rater Reliability scores for observed agreement among group of 5 annotators as well as group Krippendorf's alpha}
\label{tbl:kripp}
\end{table}

As illustrated by Table \ref{tbl:kripp}, analysis on our annotator sample shows an average observed agreement of 71.74\% and a k-alpha of 0.415.  We classify our score of 0.415 as 'fair' given the aforementioned index since euphemisms are ambiguous by nature.  Future work to build upon the corpus may take a consensus coding approach to better decide on labels.

\subsubsection{Disagreement Examples}
In the examples where annotators showed disagreement, we found their supplied interpretations to be particularly useful in examining why they may have shown differences in labeling. We attempted to identify several, consistent cases for disagreement in these examples:

\begin{enumerate}
  \item \textbf{Varying interpretations.} Annotators sometimes differed significantly in what they deemed to be the meaning of a PET, even given context. The PET "freedom fighter", for example, might be interpreted as "a person who fights for freedom" (literal) or someone who "uses violence to achieve political goals" (euphemistic). PETs interpreted to have more emotionally charged meanings within the context generally received a euphemistic label.
  \item \textbf{The use of a commonly accepted term (CAT).} Annotators tended to disagree when the PET in question was a commonly accepted term (CAT) in a particular domain (e.g., medical, journalism) or community (e.g., the disability community, LGBT+). As an example, the PET "venereal disease" can be seen as an alternative to "sexually transmitted disease" (a euphemistic usage) or simply as a CAT in the medical domain, in which case the usage is objective (a literal usage). Generally, it seems CATs could be viewed as non-euphemistic because they are the "default" term, but also as formalisms or categories in some contexts used to avoid an impolite alternative or undesired specification (euphemistic usage). The identification of CATs as a reason for ambiguity and disagreement in this task could be significant for euphemism research, as they can identified fairly clearly, and marked as needing special attention.  In this sense, we consider CATs to be fuzzy PETs since depending on the hearer's interpretation they may or may not label the term as euphemistic.
  \item \textbf{Similar interpretations.} Examples where the interpretations were nearly the same, but had varying labels, could indicate disagreement of something outside of the context. Examples include texts with PETs like "slim" and "overweight", which sometimes had disagreement despite being unanimously interpreted as "skinny" and "fat". Annotators' judgments about the nuance of these terms, or even speakers' intent, could have led to disagreement; i.e., if the use of the nuance is deliberate, the PET may be literal, but this could depend on the speaker's intent.
\end{enumerate}

For these cases, there appears to be an inherent ambiguity in the classification task, which points to ambiguity in judgments about euphemisms as a whole. Factors such as varying interpretations, the use of CATs, and subjective judgments about speaker intent may all contribute to disagreement in human interpretations of PETs. (More examples of each case can be found in Appendix E).

\section{Conclusion}
In this paper we described the creation of a new corpus of euphemistic and non-euphemistic usages of Potentially Euphemistic Terms (PETs).  We performed two experiments: 1) Sentiment Analysis with a roBERTa-base model to confirm our assumptions about how euphemisms are used to soften language, and 2) conducted a survey and observe some cases of disagreement when using euphemisms.  Our contributions were made in an effort to further along research done in automatic euphemism detection, identification and generation for a variety of NLP applications.

\section*{Acknowledgements}
We thank our annotators Raz Besaleli, Kira Horiuchi, Kelly Ortega, and Kenna Reagan for their time and attention to our corpus annotation task as well as Brad McNamee and Avery Field for their contributions to our PETs dataset.  This material is based upon work supported by the National Science Foundation under Grant No.~1704113.

\section*{References}
\bibliographystyle{lrec}
\bibliography{lrec2020W-xample-kc}


\newpage
\onecolumn
\appendix
\section{List of 71 PETs only used Euphemistically}
Below are the 71 PETs that were found to only ever be used in the euphemistic sense.  Counts for the numbers of examples per PET are provided as well.

\begin{center}
\begin{longtable}{|l|c|}
\caption{List of PETS used only euphemistically with counts.} \label{tab:long} \\

\hline \multicolumn{1}{|c|}{\textbf{PET} \emph{always euph}} & \multicolumn{1}{c|}{\textbf{No. of Euphemistic Examples}}\\ \hline 
\endfirsthead

\multicolumn{2}{c}%
{{\bfseries \tablename\ \thetable{} -- continued from previous page}} \\
\hline \multicolumn{1}{|c|}{\textbf{PET} \emph{always euph}} & \multicolumn{1}{c|}{\textbf{No. of Euphemistic Examples}}\\ \hline 
\endhead

\hline 
\multicolumn{2}{|r|}{{Continued on next page}} \\ \hline
\endfoot

\hline \hline
\endlastfoot
able-bodied                       & 7                           \\
adult beverage                    & 1                           \\
advanced age                      & 19                          \\
armed conflict                    & 20                          \\
birds and the bees                & 7                           \\
broken home                       & 1                           \\
capital punishment                & 19                          \\
comfort women                     & 3                           \\
correctional facility             & 18                          \\
custodians                        & 2                           \\
dearly departed                   & 3                           \\
deceased                          & 20                          \\
detainee                          & 20                          \\
detention camp                    & 10                          \\
developed/ing country             & 19                          \\
developmentally disabled          & 2                           \\
differently-abled                 & 2                           \\
drinking problem                  & 7                           \\
droppings                         & 18                          \\
economical with the truth         & 1                           \\
elderly                           & 20                          \\
enhanced interrogation techniques & 6                           \\
ethnic cleansing                  & 20                          \\
fatality                          & 20                          \\
freedom fighter                   & 20                          \\
full figured                      & 1                           \\
global south                      & 8                           \\
golden years                      & 17                          \\
hearing impaired                  & 2                           \\
homemaker                         & 19                          \\
income inequality                 & 20                          \\
indigent                          & 18                          \\
inebriated                        & 16                          \\
inner city                        & 20                          \\
latrine                           & 3                           \\
lavatory                          & 7                           \\
less fortunate                    & 20                          \\
lose {[}pro{]} lunch              & 4                           \\
low-income                        & 20                          \\
make love                         & 11                          \\
mentally challenged               & 17                          \\
mentally disabled                 & 11                          \\
mistruth                          & 4                           \\
negative cash flow                & 3                           \\
pass gas                          & 1                           \\
people/persons of color           & 20                          \\
physically challenged             & 1                           \\
plus-sized                        & 2                           \\
portly                            & 7                           \\
pre-owned                         & 7                           \\
pregnancy termination             & 4                           \\
pro-choice                        & 20                          \\
pro-life                          & 20                          \\
psychiatric hospital              & 11                          \\
rear end                          & 10                          \\
running behind                    & 1                           \\
same sex                          & 4                           \\
sanitation worker                 & 20                          \\
senior citizen                    & 20                          \\
sex worker                        & 20                          \\
street person                     & 3                           \\
substance abuse                   & 20                          \\
substance abuser                  & 11                          \\
targeted killing                  & 11                          \\
time of the month                 & 5                           \\
tinkle                            & 2                           \\
under the weather                 & 1                           \\
underprivileged                   & 11                          \\
undocumented immigrant            & 20                          \\
undocumented workers              & 13                          \\
venereal disease                  & 6                          

\end{longtable}
\end{center}
\newpage

\section{List of 58 PETs used Euphemistically and Literally}
Below are the 58 PETs that were found to be used in both the euphemistic and literal sense.  Counts for the numbers of sentence examples in our corpus per PET are provided as well.

\begin{center}
\begin{longtable}{|l|c|c|}
\caption{List of PETS used literally and euphemistically with counts.} \label{tab:long} \\

\hline \multicolumn{1}{|c|}{\textbf{PET} \emph{sometimes euph}} & \multicolumn{1}{c|}{\textbf{No. of Euphemistic Examples}} & \multicolumn{1}{c|}{\textbf{No. of Literal Examples}} \\ \hline 
\endfirsthead

\multicolumn{3}{c}%
{{\bfseries \tablename\ \thetable{} -- continued from previous page}} \\
\hline \multicolumn{1}{|c|}{\textbf{PET} \emph{sometimes euph}} & \multicolumn{1}{c|}{\textbf{No. of Euphemistic Examples}} & \multicolumn{1}{c|}{\textbf{No. of Literal Examples}} \\ \hline 
\endhead

\hline \multicolumn{3}{|r|}{{Continued on next page}} \\ \hline
\endfoot

\hline \hline
\endlastfoot

a certain age     & 11 & 11 \\
accident          & 26 & 6  \\
aging             & 30 & 30 \\
between jobs      & 7  & 7  \\
chest             & 10 & 10 \\
collateral damage & 26 & 26 \\
custodian         & 6  & 6  \\
demise            & 28 & 28 \\
deprived          & 2  & 2  \\
disabled          & 30 & 30 \\
disadvantaged     & 14 & 14 \\
dismissed         & 13 & 13 \\
downsize          & 2  & 2  \\
economical        & 14 & 14 \\
expecting         & 23 & 23 \\
experienced       & 2  & 4  \\
exterminate       & 15 & 15 \\
getting clean     & 2  & 2  \\
gluteus maximus   & 1  & 1  \\
go all the way    & 5  & 5  \\
got clean         & 1  & 1  \\
intoxicated       & 16 & 16 \\
invalid           & 3  & 3  \\
late              & 30 & 30 \\
lay off           & 18 & 18 \\
let {[}pro{]} go  & 7  & 7  \\
let go of         & 5  & 5  \\
long sleep        & 1  & 1  \\
mixed up          & 11 & 11 \\
neutralize        & 8  & 8  \\
oldest profession & 1  & 1  \\
outlived {[}pro{]} usefulness & 2                                    & 2                                \\
outspoken         & 3  & 3  \\
over the hill     & 6  & 6  \\
overweight        & 19 & 19 \\
pass away         & 18 & 18 \\
pass on           & 6  & 6  \\
perish            & 20 & 20 \\
plump             & 10 & 10 \\
put to sleep      & 7  & 3  \\
regime change     & 8  & 6  \\
same-sex          & 8  & 8  \\
seasoned          & 2  & 2  \\
seeing someone/each other     & 2                                    & 2                                \\
sleep around      & 1  & 1  \\
sleep with        & 6  & 6  \\
slim              & 11 & 13 \\
sober             & 11 & 11 \\
special needs     & 13 & 13 \\
stout             & 6  & 6  \\
to go to heaven   & 1  & 1  \\
troubled          & 15 & 15 \\
underdeveloped    & 13 & 13 \\
wealthy           & 5  & 5  \\
weed              & 30 & 30 \\
well off          & 11 & 11 \\
went to heaven    & 1  & 1  \\
with child        & 2  & 2  \\

\end{longtable}
\end{center}
\newpage

\section{Relative changes in sentiment score per PET}
Shown below is a sample of the relative changes in sentiment and offensiveness scores produced by roBERTa models after substitution of literal meanings, grouped and averaged by PET. For the full list, see the GitHub page.
\newline

\begin{filecontents*}{sentiment_diffs_by_type_sample.csv}
type,neu change,pos change,neg change,off change,n-off change
a certain age,-0.027845,-0.017681,0.769987,0.068562,-0.016036
able-bodied,0.061313,0.135860,-0.056808,-0.100490,0.039479
accident,-0.062816,-0.156983,1.062615,0.543362,-0.142711
adult beverages,0.117982,-0.043823,0.199893,0.113861,-0.009353
advanced age,-0.026644,-0.177761,0.358272,0.094902,-0.018991
aging,-0.045431,-0.053698,0.179980,0.092581,-0.019898
armed conflict,-0.025140,0.008283,0.031437,0.035220,-0.007695
between jobs,-0.080394,-0.223505,2.631469,0.280354,-0.021428
birds and the bees,-0.023246,-0.373782,0.751403,1.847629,-0.297745
broken home,-0.036726,-0.066531,0.027881,0.004441,-0.005638
capital punishment,-0.056940,-0.110768,0.141521,0.132765,-0.048826
chest,0.001200,-0.018163,0.010530,0.200816,-0.087369
collateral damage,-0.272639,-0.369028,0.320184,0.438191,-0.104633
comfort women,-0.129123,-0.142995,0.043196,0.089508,-0.082556
correctional facility,-0.047930,-0.064335,0.213246,0.087022,-0.023350
custodian,-0.022650,-0.058305,0.178856,0.213095,-0.024849
custodian ,-0.003269,0.079685,-0.110462,0.010495,-0.000924
dearly departed,0.019440,-0.343311,1.152702,0.490740,-0.051384
deceased,-0.000788,-0.079376,0.391194,0.188400,-0.035379
demise,-0.012091,-0.019827,0.021183,0.138516,-0.041601
deprived,-0.015199,0.060084,0.011545,0.212837,-0.040061
detainee,-0.019231,0.005419,0.026933,0.056855,-0.017363
detention camp,0.055365,0.158105,-0.001760,-0.002855,-0.008876
developing/ed country,-0.005165,-0.188186,0.424494,0.264971,-0.034276
developmentally disabled,-0.342488,-0.395191,0.636818,0.905488,-0.409589
differently-abled,-0.059591,-0.008284,0.020493,0.135978,-0.041278
disabled,0.003972,-0.061131,0.140886,0.107509,-0.022856
disadvantaged,-0.005827,-0.122688,0.449368,0.195209,-0.033192
dismissed,0.138839,0.115313,-0.084370,0.157943,-0.038344
downsize,-0.064027,-0.181702,0.048146,0.333272,-0.052051
drinking problem,0.025495,0.073228,0.018369,-0.061759,0.056646
droppings,-0.227458,-0.279421,0.303311,0.355247,-0.196141
economical,-0.097051,0.194375,0.081696,0.194043,-0.031545
economical with the truth,-0.454708,-0.504673,0.490267,1.576375,-0.248274
elderly,0.046045,0.009758,0.036105,0.031266,-0.011184
enhanced interrogation techniques,-0.162409,-0.303567,0.149646,0.230497,-0.083666
ethnic cleansing,-0.116799,-0.050420,0.060452,0.001620,0.005740
expecting,0.328630,-0.168126,1.675960,0.440075,-0.042933
experienced,-0.052473,-0.109398,0.214863,0.257064,-0.029485
exterminate,-0.202532,-0.297179,0.179729,0.206389,-0.135381
fatality,-0.062864,-0.073644,0.055863,0.100625,-0.021019
freedom fighter,-0.029893,-0.151280,0.193121,0.291315,-0.065337
full figured,-0.595011,-0.747444,2.457650,0.672804,-0.593019
getting clean,-0.074889,-0.206926,0.316258,0.010324,0.007877
global south,-0.024629,0.020819,0.035014,0.002640,0.005800
gluteus maximus,-0.004746,-0.108652,0.183191,0.808259,-0.201961
go all the way,0.153388,-0.175044,0.433929,1.910859,-0.228310
golden years,0.070800,-0.162567,0.208371,0.289411,-0.036933
got clean,-0.191045,-0.363818,0.531799,1.568274,-0.262431
hearing impaired,-0.141219,-0.060333,0.080978,0.467513,-0.232416
homemaker,0.009685,-0.063941,0.136986,0.184696,-0.019186
\end{filecontents*}
\csvautotabular{sentiment_diffs_by_type_sample.csv}

\newpage

\section{Annotation Task Instructions}
\includegraphics[scale=.85, page = 1]{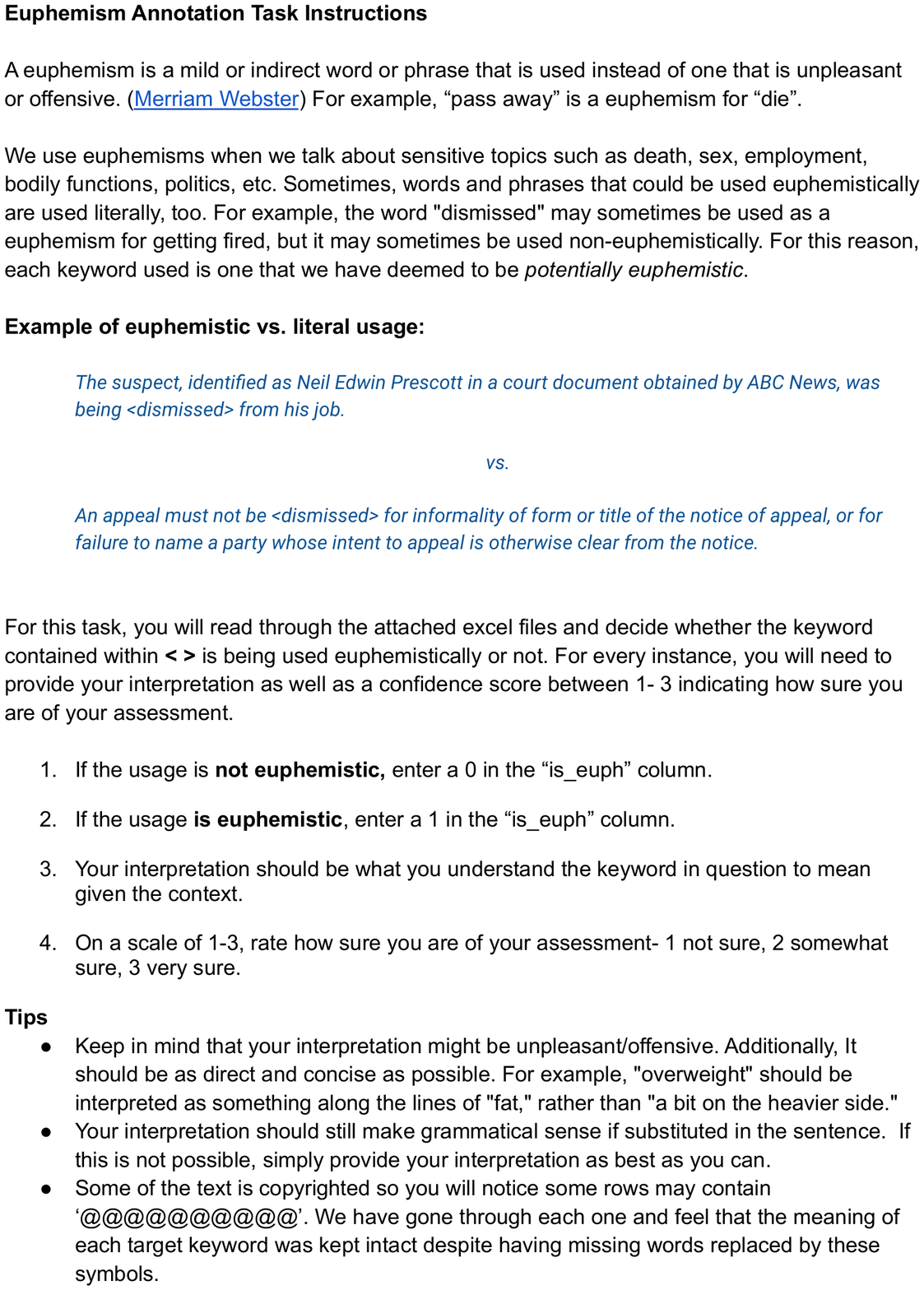}

\section{Examples of annotator disagreement, organized by possible causes}
Below are example texts where annotators differed in their labels, and had significantly varying interpretations of the PET.
\begin{tabular}{|P{1.25cm}|P{4.75cm}|P{2.25cm}|P{2.25cm}|P{2.25cm}|P{2.25cm}|}
\hline
\textbf{PET}& \textbf{text}& \textbf{annotator1} & \textbf{annotator2} & \textbf{annotator3} & \textbf{annotator4}\\ 
\hline
freedom fighter & 
The Palestinian woman is unlike @ @ @ @ @ @ @ @ @ @ be the daughter, wife, sister or mother of the prisoners, the dead, the injured. She is a stone thrower, a \textless freedom fighter\textgreater. She is an inspiration for every woman as she smiles, works hard, revolts, rebels and dreams despite all what all difficulties she does through. & 
\textbf{Label:} 0 
\newline
\textbf{Interpretation:} someone with strong beliefs and actions &
\textbf{Label:} 0 
\newline
\textbf{Interpretation:} actively opposing suffering under occupation & 
\textbf{Label:} 0 
\newline
\textbf{Interpretation:} a person who fights for freedom &
\textbf{Label:} 1 
\newline
\textbf{Interpretation:} uses violence to achieve political goals
\\
\hline
collateral damage & 
This is very very bad; that we have a CIC that will relieve field commanders for doing the right thing. I guess we are supposed to consider our Benghazi dead as \textless collateral damage \textgreater to BO's foreign policy.& 
\textbf{Label:} 1 
\newline
\textbf{Interpretation:} injuries that were not intended &
\textbf{Label:} 0 
\newline
\textbf{Interpretation:} a talking point & 
\textbf{Label:} 1 
\newline
\textbf{Interpretation:} unintended deaths due to war &
\textbf{Label:} 0 
\newline
\textbf{Interpretation:} damage as consequence
\\
\hline
\end{tabular}
\newline
\newline
Below are example texts where annotators differed in their labels, and the PET in question could be considered a CAT.
\begin{tabular}{|P{1.25cm}|P{4.75cm}|P{2.25cm}|P{2.25cm}|P{2.25cm}|P{2.25cm}|}
\hline
\textbf{PET}& \textbf{text}& \textbf{annotator1} & \textbf{annotator2} & \textbf{annotator3} & \textbf{annotator4}\\ 
\hline
disabled & 
Your life testifies to the fact that life with disability can be abundantly fulfilling. Here in Nigeria, my organization AAD INITIATIVE is working toward that-fulfilling life for the \textless disabled \textgreater, and hopefully we shall get there. & 
\textbf{Label:} 0 
\newline
\textbf{Interpretation:} disabled &
\textbf{Label:} 1 
\newline
\textbf{Interpretation:} those who have a disability, patronizing & 
\textbf{Label:} 0 
\newline
\textbf{Interpretation:} physically or mentally impaired &
\textbf{Label:} 0 
\newline
\textbf{Interpretation:} broad term for people with disability
\\
\hline
pro-choice & 
Since there is no agreement, the \textless Pro-Choice \textgreater position (not a belief) understands reality. The decision to give birth is exclusively a woman's-- whether or not it is legal.& 
\textbf{Label:} 1 
\newline
\textbf{Interpretation:} pro abortion rights &
\textbf{Label:} 0 
\newline
\textbf{Interpretation:} commonly accepted name of a sociopolitical movement & 
\textbf{Label:} 1 
\newline
\textbf{Interpretation:} pro-abortion &
\textbf{Label:} 0 
\newline
\textbf{Interpretation:} a term that represents people who know that women should make choices for their body
\\
\hline
\end{tabular}
\newline
Below are example texts where annotators differed in their labels, despite supplying very similar interpretations.
\begin{tabular}{|P{1.25cm}|P{4.75cm}|P{2.25cm}|P{2.25cm}|P{2.25cm}|P{2.25cm}|}
\hline
\textbf{PET}& \textbf{text}& \textbf{annotator1} & \textbf{annotator2} & \textbf{annotator3} & \textbf{annotator4}\\ 
\hline
slim & 
It has become much more common over the past few years because people are eating too much fat and not enough starch and fiber. People who eat a diet based on low-fat, unrefined plant foods stay naturally \textless slim\textgreater. & 
\textbf{Label:} 0 
\newline
\textbf{Interpretation:} skinny &
\textbf{Label:} 0
\newline
\textbf{Interpretation:} skinny & 
\textbf{Label:} 1
\newline
\textbf{Interpretation:} skinny &
\textbf{Label:} 0 
\newline
\textbf{Interpretation:} skinny - physical description
\\
\hline
able-\newline bodied & 
[...] The men with the poster could claim that they were not simply begging, but were doing what was expected of a parent who could not afford to care for a sick child. Some \textless able-bodied \textgreater beggars accompanied a blind or crippled-parent as a means to solicit sympathy from donors [...] & 
\textbf{Label:} 0
\newline
\textbf{Interpretation:} not disabled &
\textbf{Label:} 0 
\newline
\textbf{Interpretation:} opposite of disabled & 
\textbf{Label:} 1 
\newline
\textbf{Interpretation:} not disabled &
\textbf{Label:} 1 
\newline
\textbf{Interpretation:} not disabled
\\
\hline
\end{tabular}

\end{document}